\documentclass[conference]{IEEEtran}
\IEEEoverridecommandlockouts
\usepackage{cite}
\usepackage{amsmath,amssymb,amsfonts}
\usepackage{graphicx, caption, subcaption}
\usepackage{textcomp}
\usepackage[table]{xcolor}
\usepackage{algorithm,algpseudocode}
\usepackage{hyperref}
\hypersetup{
     colorlinks = true,
     citecolor = green,
     }
\usepackage{lipsum}
\usepackage{multirow}
\usepackage{booktabs}
\usepackage[usestackEOL]{stackengine}
\usepackage{fancyhdr}

\usepackage{ifthen}

\usepackage[utf8]{inputenc}
\usepackage[T1]{fontenc}
\usepackage{pifont}
\DeclareUnicodeCharacter{2716}{\ding{54}}
\usepackage{microtype}

\def\BibTeX{{\rm B\kern-.05em{\sc i\kern-.025em b}\kern-.08em
    T\kern-.1667em\lower.7ex\hbox{E}\kern-.125emX}}
\begin{document}

\newboolean{show_review_colors}
\setboolean{show_review_colors}{true}
% \setboolean{show_review_colors}{false}

\ifthenelse{\boolean{show_review_colors}}{%
    \newcommand{\wrong}[1]{{\color{red}{{#1}}}}
    \newcommand{\review}[1]{{\color{blue}{{#1}}}}
    \newcommand{\elaborate}[1]{{\color{green}{{#1}}}}
}{%
    \newcommand{\wrong}[1]{{{{#1}}}}
    \newcommand{\review}[1]{{{{#1}}}}
    \newcommand{\elaborate}[1]{{{{#1}}}}
}

% how to architect safety around a voodoo-infused software systems by systematically evaluating the performance of two supervisors using a large-scale AD dataset for the first time in this unique configuration

\title{Evaluation of Out-of-Distribution Detection Performance on Autonomous Driving Datasets}

\author{
\IEEEauthorblockN{Jens Henriksson\IEEEauthorrefmark{1}, 
Christian Berger\IEEEauthorrefmark{2}, 
Stig Ursing\IEEEauthorrefmark{1}, 
Markus Borg\IEEEauthorrefmark{3}
%, Kasper Socha\IEEEauthorrefmark{3}
}\\
\IEEEauthorblockA{\IEEEauthorrefmark{1}Semcon AB, Gothenburg, Sweden, Email: \{jens.henriksson, stig.ursing\}@semcon.com}

\IEEEauthorblockA{\IEEEauthorrefmark{2}University of Gothenburg, Sweden, Email: christian.berger@gu.se}

\IEEEauthorblockA{\IEEEauthorrefmark{3}Lund University, Lund, Sweden, Email: markus.borg@cs.lth.se}

%\IEEEauthorblockA{\IEEEauthorrefmark{3}RISE Research Institutes of Sweden AB, Lund and Gothenburg, Sweden, Email: \{markus, cristofer\}@ri.se}
%\IEEEauthorblockA{\IEEEauthorrefmark{4}Machine Learning and AI Center of Excellence, Volvo Cars, Gothenburg, Sweden, Email: lars.tornberg@volvocars.com}
%\IEEEauthorblockA{\IEEEauthorrefmark{5}QRTech AB, Gothenburg, Sweden, Email: sankar.sathyamoorthy@qrtech.se}

}

\maketitle

% \thispagestyle{plain}
% \pagestyle{plain}

% Adding header and page number to postprint version
\pagestyle{fancy}
\thispagestyle{fancy}
\fancyhf{}
\chead{\textcolor{red}{\textbf{Preprint} -- Accepted to \textit{2023 IEEE International Conference On Artificial Intelligence Testing (AITest)}}}
%\chead{\textbf{Preprint} -- Accepted to REFSQ 2023}
\cfoot{\thepage}
\renewcommand{\headrulewidth}{0pt}
%---------------------------------------

\begin{abstract}
Safety measures need to be systemically investigated to what extent they evaluate the intended performance of Deep Neural Networks (DNNs) for critical applications. Due to a lack of verification methods for high-dimensional DNNs, a trade-off is needed between accepted performance and handling of out-of-distribution (OOD) samples. 

This work evaluates rejecting outputs from semantic segmentation DNNs by applying a Mahalanobis distance (MD) based on the most probable class-conditional Gaussian distribution for the predicted class as an OOD score. The evaluation follows three DNNs trained on the Cityscapes dataset and tested on four automotive datasets and finds that classification risk can drastically be reduced at the cost of pixel coverage, even when applied on unseen datasets. The applicability of our findings will support legitimizing safety measures and motivate their usage when arguing for safe usage of DNNs in automotive perception.

\end{abstract}


%%%%%%%%%%%%%%%%%%%%%%%%%%%%%%%%%%%%%%%%%%
%%%%%%%%%%%%%%%%%%%%%%%%%%%%%%%%%%%%%%%%%%
\section{Introduction}\label{sec:intro}
The power of data-driven algorithms such as deep neural networks (DNNs) has enabled a new era of algorithms to conduct challenging tasks in several different domains. For autonomous driving, perception has seen an increased performance by incorporating DNNs to handle complex tasks. One of the perception tasks is semantic segmentation, the task of classifying each pixel into a semantic category. The benchmark dataset Cityscapes has had top contenders with different DNN architectures for the past years~\cite{Cordts2016TheUnderstanding}. 

Unfortunately, DNNs are inherently difficult to analyze~\cite{Borg2019SafelyIndustry}. The automotive industry has over the years developed and incorporated significant standards such as ISO~26262~\cite{ISO26262} to emphasize the importance of risk reduction through a multitude of methods that all aim to simplify the verification and validation of software items. In addition, studies of the standard highlighted that most methods are not applicable or useful for DNN development~\cite{Salay2017}, which was partially a reason why ISO~21448 SOTIF~\cite{ISO21448SOTIF} was developed and additional standards are on the horizon, e.g., ISO/PAS~8800~\cite{ISO8800}. We have previously demonstrated how to develop a safety argumentation for a DNN-based perception system in the SOTIF context~\cite{borg2022ergo}.

% Even though DNNs have shown superior results in this task, it comes with an inherent challenge in verifying and validating that the correct behavior is maintained on data that contains slight differences compared to what the model is trained on. In addition, as an input image can have a size containing hundreds of thousands of pixels, small variations in a fraction of these parameters can render the output from the model completely different. 

% The field of generative models utilize this phenomenon to construct samples that can fool the DNN to produce false detections or classifications, typically by adding a small variation to the input image~\cite{GoodfellowGenerativeNets}. While these specific samples are hand-crafted, similar real-life samples are common in unseen driving scenarios.  

\begin{figure}
    \centering
    \includegraphics[width=0.49\linewidth]{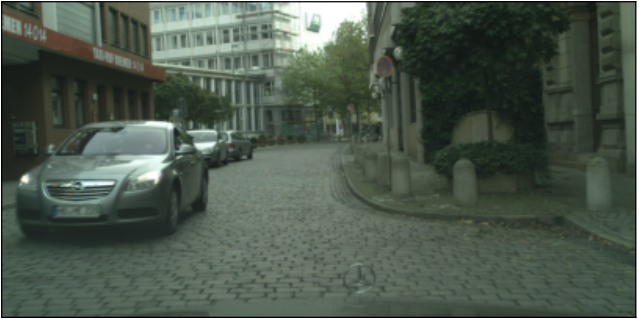}
    \includegraphics[width=0.49\linewidth]{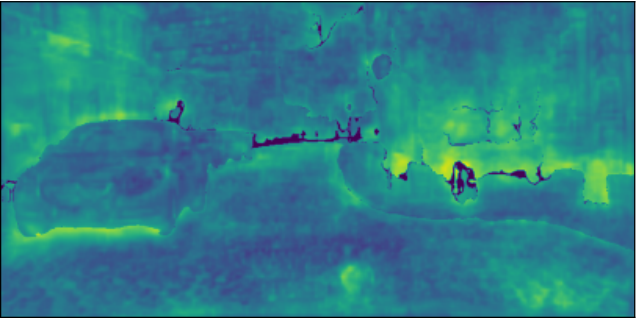}
    \caption{A sample of the class conditional Mahalanobis distance on a training image. Brighter colors refer to larger distances.}
    \label{fig:intro_md_vis}
\end{figure}

One of the issues for the verification of perception systems is the difficulty of constructing a proper specification for the task at hand. A class cannot be completely specified in the complex input space; for example, no complete definition of what constitutes a \textit{pedestrian} can be defined in an image due to its high-dimensional space. In addition, high-dimensional data suffers from several issues regarding anomaly detection, including vanishing distance metrics and irrelevant parameters~\cite{Zimek2012AData}. Combining the difficulties in the verification of image-based DNNs with images' high dimensionality, guaranteeing the models' prediction performance becomes an infeasible task. Instead, safety measures that reduce misclassification risk in safety critical applications are needed~\cite{henriksson2023refsq}. 

% Instead, researchers have accepted that there will be a trade-off between coverage and risk of misclassification of objects or scenes. 

% \review{For high-dimensional data, Zimek et al. identified several issues regarding anomaly detection. These issues include vanishing distance metrics, irrelevant parameters, i.e., a majority of pixels being unused and challenges with comparing or interpreting scores between samples as they can be composed by different sub-spaces and thus, being incomparable.}

% Another issue from a formal verification point-of-view, is the fact that it is infeasible to come up with a formal definition of what constitutes a specific object class in an input image. There is no definition of what constitutes a pedestrian in an automotive dataset. A detailed review of verification and validation techniques were conducted by Borg et al.~\cite{Borg2019SafelyIndustry}.

% \review{Combining the issues with verification and validation of DNNs partly due to the lack of class specification with the fact that high-dimensionality causes difficulties in detection of outliers yields an infeasible solution to develop camera algorithms with safety in mind. Instead, researchers have accepted that there will be a trade-off between coverage and risk of misclassification of objects or scenes. }

%A common way of demonstrating the trade-off is through studying the \textit{receiver operating characteristics}, a graphical instrument demonstrating the ability to distinguish between true positives and false positives. 

This paper showcases that a risk-coverage trade-off exists, i.e., a reduction of misclassifications can be achieved at the cost of less pixels predicted, which can be used in safety argumentation for safety critical applications. The trade-off is based on an accepted distance measure, commonly used in Out-of-distribution (OOD) detection, one method that is suggested for safety verification~\cite{Mohseni2022TaxonomyPrimer}. As OOD measure, the Mahalanobis distance (MD) is used 
as it allows a statistical distance method that compares each pixel to a class-conditional probability distribution, 
%as it allows a statistical distance comparison for each pixel to a class-conditional probability distribution,
visualized in Fig.~\ref{fig:intro_md_vis}. The measure is applied at the pixel level of three semantic segmentation DNNs trained on Cityscapes~\cite{Cordts2016TheUnderstanding}, and then further evaluated on three additional datasets: BDD100K~\cite{Yu2020BDD100K:Learning}, A2D2~\cite{GeyerA2D2:Dataset}, and KITTI-360~\cite{Liao2022KITTI-360:3D}, with the following research questions studied:

% for the inclusion of DNNs in safety critical applications by addressing the following research questions:

% are used to create varied testing sets that is distributed over cities in the US, Germany, Switzerland, and Israel. We decide on one distance-based outlier score, namely the Mahalanobis distance (MD) as a statistical method that allows each pixel to be compared to a class-conditional probability distribution~\cite{Lee2018}. More specifically, this paper studies if the MD can reduce the risk of misclassifications from a trained DNN by addressing the following research questions:

\begin{itemize}
    \item[RQ1] To what extent can Mahalanobis distance (MD) be used to reduce misclassification at the trade-off of covered pixels?
    \item[RQ2] To what degree does the trade-off vary when applied to samples outside of the training set of the model?
    % \item[RQ1] To what extent can the Mahalanobis distance (MD) be used to separate between inlier and outlier images, i.e., if the sample is from a city that is in the training set or not? 
\end{itemize}

The remainder of this paper is organized as follows: Section~\ref{sec:rw} discusses related work within safety and outlier detection, Section~\ref{sec:method} introduces the methodology and reasoning behind the experiments conducted. In Section~\ref{sec:results}, the results are presented which are then further discussed in Section~\ref{sec:fw}. 

%%%%%%%%%%%%%%%%%%%%%%%%%%%%%%%%%%%%%%%%%%
%%%%%%%%%%%%%%%%%%%%%%%%%%%%%%%%%%%%%%%%%%
\section{Related Work}\label{sec:rw}
%Markus to write a related work section, inspired by: https://tsigalko18.github.io/assets/pdf/2022-Stocco-ASE.pdf

Several studies on automotive software engineering explore automated online recognition of unfamiliar input for perception systems, i.e., OOD scenarios. Detecting unknown and uncertain conditions is vital for the development of safe autonomous vehicles. The software testing community suggests several ways to support the verification and validation of perception systems and their OOD detection capabilities. In this section, we discuss OOD detection in general and the most related work on online testing and monitoring of automotive perception systems.

\begin{figure*}[!t]
    \centering
    \includegraphics[width=0.215\textwidth]{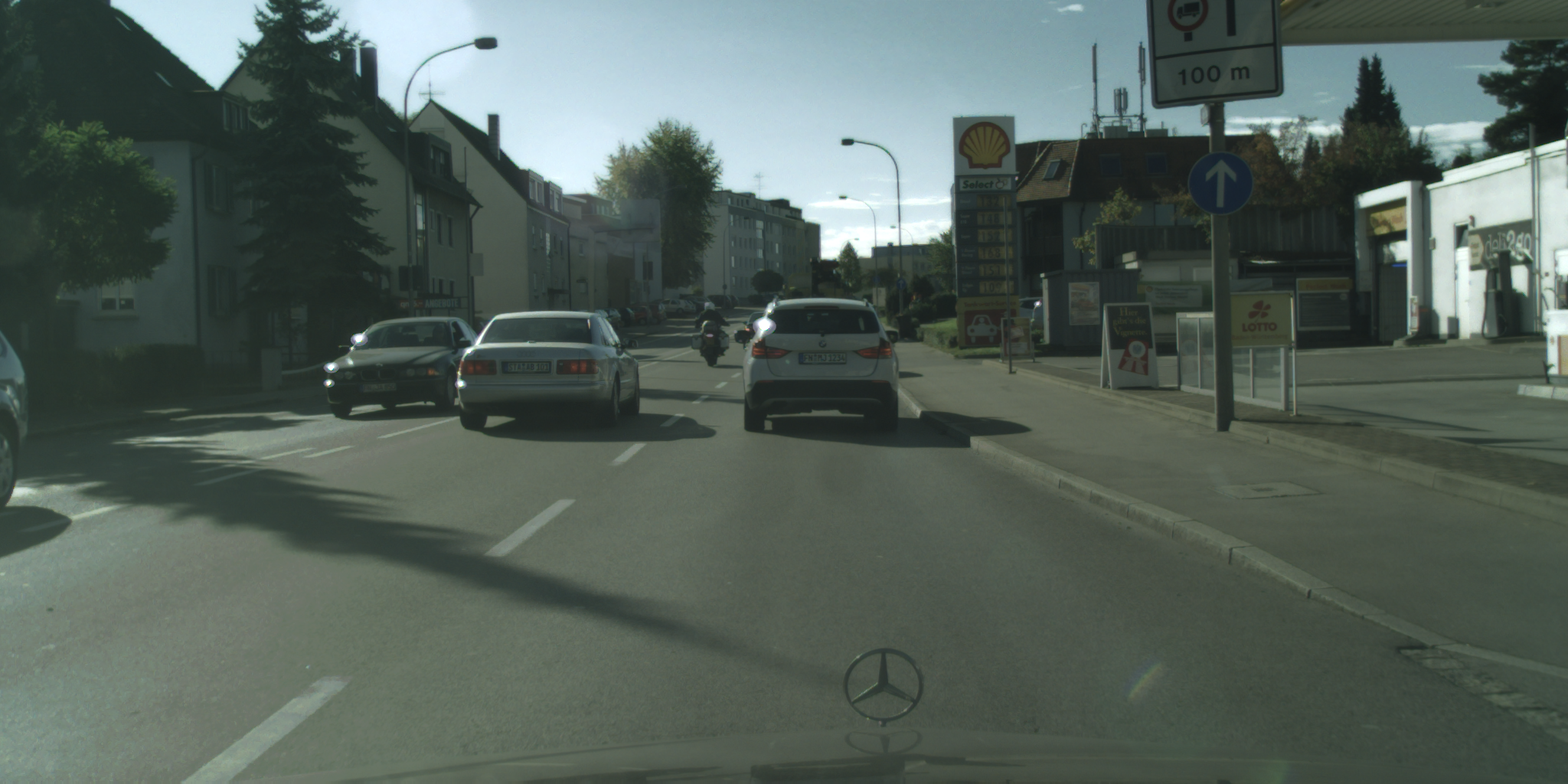}
    \includegraphics[width=0.1911\textwidth]{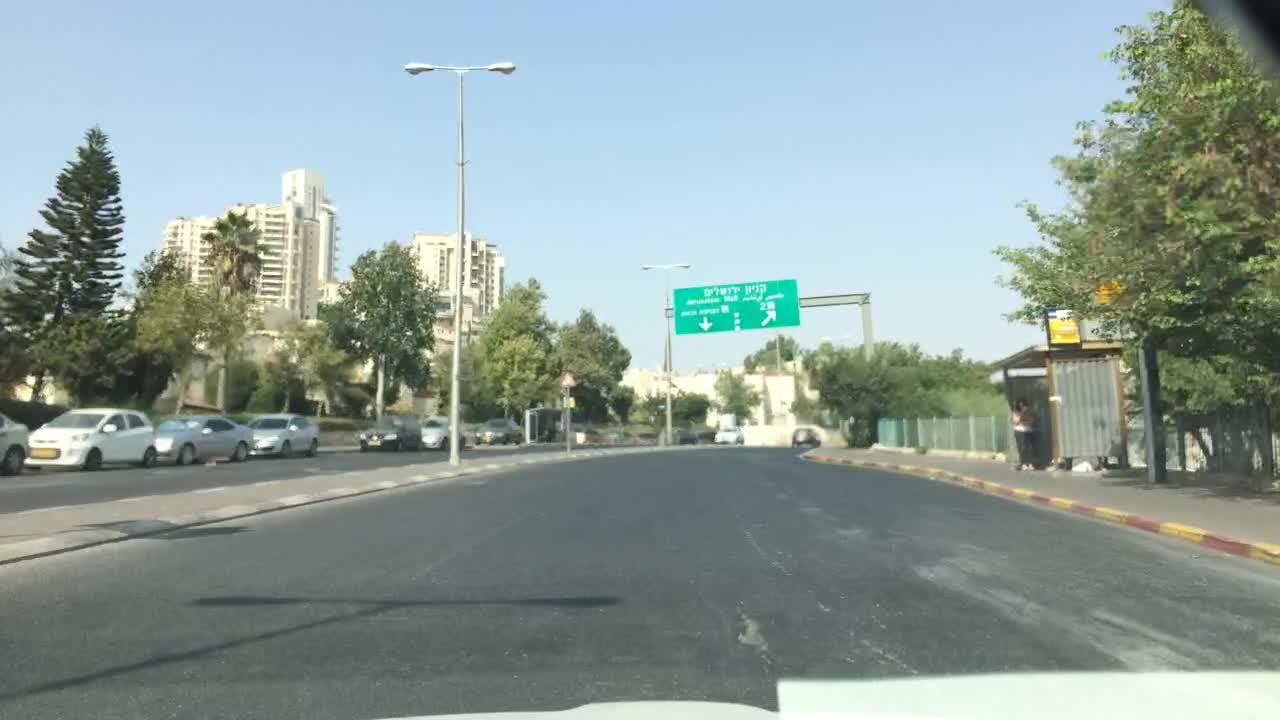}
    \includegraphics[width=0.402\textwidth]{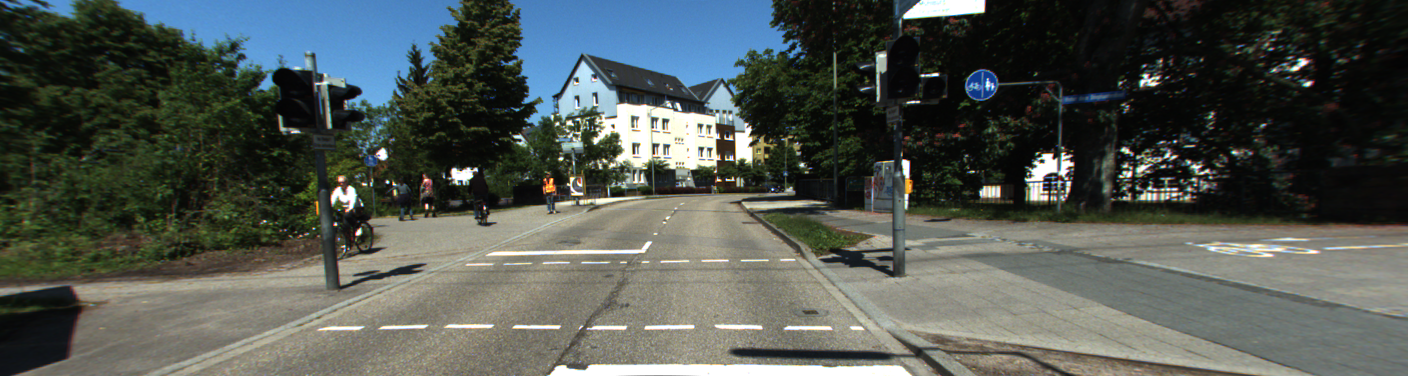} \includegraphics[width=0.17\textwidth]{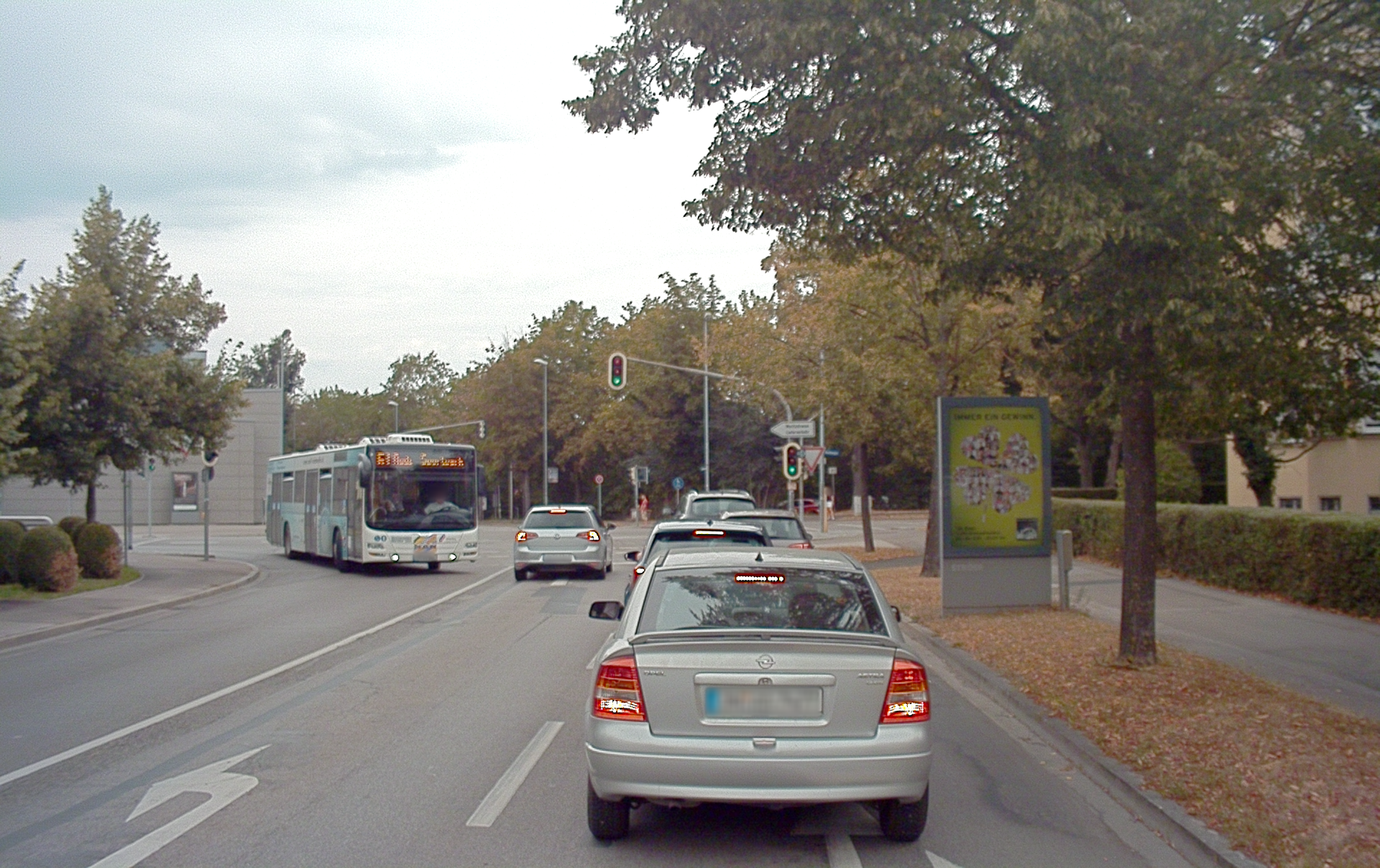}
    \caption{Sample images for the four datasets. From left: Cityscapes, BDD100K, KITTI-360, and A2D2. The images maintain their original aspect ratio.}
    \label{fig:test_set_images}
\end{figure*}

\subsection{General OOD Detection}
A recent survey on generalized OOD detection was conducted by Yang et al.~\cite{Yang2021GeneralizedSurvey}. In their survey, they describe the differences between anomaly detection, novelty detection, open set recognition and OOD detection. Furthermore, their survey also distinguishes methods into classification-based, density-based, distance-based and reconstruction-based methods. This paper falls into the distance-based OOD detection category.

One of the first adoptions of OOD detection on imagery data was conducted by Hendrycks et al.~\cite{Hendrycks2017}. Their method utilized the most probable prediction of the classifier's posterior distribution as a distance metric. 

Zhang et al.~\cite{zhang2021out} proposed a notion of relative activation and deactivation to interpret the inference behaviour of a DNN. Using this notion, the authors developed an OOD detection algorithm and demonstrated promising results on eight standard computer vision benchmarks. However, the application of the algorithm is non-trivial as it requires an understanding of the DNN on the level of layers.

Xiao et al.~\cite{xiao2021self} presented SelfChecker, a DNN monitoring system that triggers an alarm when features of the internal DNN layers are inconsistent with the final prediction. SelfChecker is inspired by the finding from Kaya et al. that a DNN can find the reach prediction before the final layer but then change it in the output layer~\cite{kaya2019shallow}. The concept of self-checking involves monitoring whether the internal layers and the output layer are inconsistent. This approach can be used for OOD detection but again requires a layer-level understanding of the DNN.

Lee et al.~\cite{Lee2018} constructed an OOD-detection method that utilized a class-conditional Gaussian distribution constructed from the training set, which was then used as the basis to receive the Mahalanobis distance for a given input sample. Their work outperforms the method from Liang et al.~\cite{Liang2018}, which utilizes the fact that input perturbations harm inlier data samples more than outliers. The novel and original contribution from our paper is about extending this method and applying it to more complex images as well as on a per-pixel level rather than per-image basis.

We have previously also proposed a framework with seven metrics to support systematic evaluations of candidate OOD detectors~\cite{Henriksson2019TowardsSupervisors}. In this paper, we modify the coverage breakpoint graph to visualize the trade-off between risk and model coverage and put it into context of applicability to safety requirements for DNNs. 

% Similar to detection of undesired samples, from the adversarial detection domain, Chen et al.~\cite{Chen2017ReabsNet:Examples} introduces an extension of a regular DNN with a guardian network that first identifies adversarials, then proceed with trying to construct a normal sample from the adversarial that can be proceeded through the network. 

\subsection{OOD Detection for Automotive Perception Systems}
OOD detection is a popular research topic in the automotive domain. The typical idea is to complement the DNN with a supervisor that can detect anomalies, which could indicate corner cases or when the vehicle leaves the operational design domain (ODD). An online supervisor can be used to build a safety envelope over a DNN, also known as a safety cage architecture~\cite{heckemann2011safe}. Bogdoll and Nitsche recently published an overview of different OOD detectors proposed for automotive perception systems~\cite{bogdoll2022anomaly}.

Several other researchers have addressed OOD detectors. Zhang et al.~presented pioneering work that compared distances between single input images and the training set~\cite{zhang2018deeproad}. Similar to our previous safety mechanism~\cite{borg2022ergo}, Hussain et al.~relied on a variational autoencoder (VAE)~\cite{hussain2022deepguard}. Hell et al.~also used a VAE but also showed that two alternatives work better in their experiments using the CARLA simulator: 1) Likelihood Regret and 2) SSD, i.e., generative modelling that uses self-supervised representation learning~\cite{hell2021monitoring}.

Stocco and colleagues have published several related studies on OOD detection for autonomous vehicles. They firstly stressed the importance of continual learning of anomaly detectors~\cite{stocco2020towards}. In their preliminary work, they demonstrated how this approach can adapt to changes while reducing the false positive rate and maintaining the original accuracy of OOD detection. The authors later refined their ideas into ``confidence-driven weighted retraining'' and provided extensive evaluations using the Udacity simulator~\cite{stocco2021confidence}. In their most recent work, they used attention maps~\cite{stocco2022thirdeye}, a popular technique in the domain of explainable artificial intelligence. The underlying idea is to turn attention maps into confidence scores and interpret uncommon attention maps as unexpected driving conditions. % Empirical evaluations in the Udacity simulator show that the solution, namely ThirdEye, can accurately detect anomalies at an early stage.

For automotive OOD detection, Oberdiek et al.~\cite{OberdiekDetectionSegmentation} demonstrated that semantic segmentation networks are suitable for OOD detection. Their experiments were conducted on automotive datasets and showed that a meta-segmentation can be used to detect unknown objects. Similarly, Di Biase et al.~\cite{BiasePixel-wiseScenes} constructed a pixel-wise anomaly detector based on uncertainty maps constructed by softmax entropy and fed that into a spatial-aware dissimilarity network.

In our previous work, we have demonstrated how OOD detection using the reconstruction error of a VAE can be used as a safety mechanism in an automotive safety case for a DNN~\cite{borg2022ergo}. We have also explored other OOD detectors~\cite{Henriksson2021Performance}, including OpenMax~\cite{Bendale2016} and ODIN~\cite{Liang2018}. OOD detection has a place in the automotive safety lifecycle as it can support evaluation of safety requirements at different stages of the development stage~\cite{henriksson2023refsq}. 

In contrast to the related work presented, this study investigates if the location of the data acquisition matters for OOD detection as well as to what extent the Mahalanobis distance method from Lee et al.~\cite{Lee2018} can act as a suitable metric when operating outside of the designed limits of the function. In addition, related articles do not consider the fraction of rejected samples, something we refer to as the risk-coverage trade-off.

% \jens{Write a concluding paragraph... two sentences... that positions our paper in light of previous work in this subsection.}

%\subsection{Automotive Software testing}

%Pei et al.~\cite{Pei2017DeepXplore:Systems} introduce a whitebox testing framework called DeepXplore, a framework that utilizes neuron coverage as a metric. The neuron coverage metric can be used to highlight samples with high activity for various behaviors through a joint optimization with gradient ascent of several DNNs. Tian et al. present DeepTest~\cite{Tian2018DeepTest:Cars} that utilizes the neuron coverage metric as a guide for test case generation of a single DNN. With this guidance, and leveraging metamorphic testing their approach can identify erroneous behavior within automotive dataset samples. Another framework for finding misclassified perturbations is presented by Huang et al.~\cite{Huang2017SafetyNetworks}. Their approach takes it one step further: to guarantee that if a perturbation exist, they will find it as they limit the search for adversarial samples. A key concept for their framework is the manipulations they add to images which are founded on real-life perturbations such as bad weather, scratches or weird angles. Finding and correcting perturbations is a challenging task by itself. However, in contrast to detection and generation of adversarials, our work studies how erroneous behavior may exist, simply by studying the same issue at different locations. 

%%%%%%%%%%%%%%%%%%%%%%%%%%%%%%%%%%%%%%%%%%
%%%%%%%%%%%%%%%%%%%%%%%%%%%%%%%%%%%%%%%%%%
\section{Methodology}\label{sec:method}
This section describes the parts needed to evaluate the risk-coverage trade-off for semantic segmentation DNNs. In short, it covers A) datasets and the selection process conducted, B) selection of models that were trained on the Cityscapes dataset, C) a detailed description of the evaluation metrics and how the risk-coverage trade-off is generated, and finally D) how to combine the previous parts to conduct the evaluation. 
% This section describes the flow of the setup to reach the evaluation of the research questions. In short, it covers 1) dataset selection that allows location-based information and which model that we are using, 2) distance-based outlier scores, and 3) evaluation criterion.

\subsection{Datasets}\label{sec:data}
A critical performance issue stems from how well DNNs can generalize their performance to data that is similar, but not part of the training iterations. Normally, generalization is estimated from the corresponding validation and test sets connected to the training set. To evaluate generalization further in this paper, a search for comparable semantically labeled automotive datasets was conducted. The criteria to be included are to contain comparable label and image dimensions to Cityscapes, as well as a way to confirm that the image is not gathered in the same city, to study to what degree models can generalize performance to different locations. 

The Cityscapes dataset provides a unique and useful distribution of their dataset by grouping their labeled images by city. No other dataset was found so far with this structure, but for some others we could instead infer the location-based information through geolocation based on GPS coordinates. In those cases where the inferred location is outside of a city, e.g., in a close-by village or country road it is still considered part of the ``city''.

% To enable the experiments in this paper, a search for comparable semantically labeled automotive images datasets was conducted. The criteria to be included is to contain comparable labels and image format to Cityscapes~\cite{Cordts2016TheUnderstanding}, location based information and be facing forward in the vehicles direction. The Cityscapes dataset provides a rather unique but useful distribution of their dataset by grouping their labeled images by city and will act as baseline with their labeling schematic. No other dataset was found with this structure, instead the location based information was inferred through geo-location based on GPS-coordinates. In those cases where the inferred location is outside of the city, e.g., in a close-by village or country road it is still considered being part of the ``city''.

The initial set of datasets to be evaluated are taken from public recommendation by Scale\footnote{\href{https://scale.com/open-av-datasets?data=Image&annotation=Semantic+Label}{scale.com}'s list of recommended semantic segmentation datasets},  as well as KITTI-360's summary of semantic segmentation datasets (cf.~Table 1 in \cite{Liao2022KITTI-360:3D}). In total, four datasets fulfilled the needs for this study: Cityscapes~\cite{Cordts2016TheUnderstanding}, KITTI-360~\cite{Liao2022KITTI-360:3D}, Audi Autonomous Driving Dataset (A2D2)~\cite{GeyerA2D2:Dataset}, and Berkeley Deep Drive (BDD100K)~\cite{Yu2020BDD100K:Learning}. 

% For KITTI-360 and A2D2, the image-label pair was provided with additional metadata containing GPS information. 

For KITTI-360 and A2D2, the image-label pair was provided with additional metadata containing GPS information. For the BDD100K dataset however, additional metadata is only available for their object detection dataset, and is not provided for the semantic segmentation part of the dataset. Furthermore, the BDD100K documentation states that due to legacy reasons, it is not guaranteed that the semantic segmentation images exist within the larger object detection dataset. Luckily, our experimental preparations showed that a large portion of the semantic segmentation images can be found in the object detection metadata by matching the unique identifiers in the two sets. This enables us to extract location information through the GPS coordinates within the metadata and extract the images that have location information. We could then sort them based on the country and city where the images were recorded. We found that a majority of images are from the west and east coasts of USA, but a small subset is from districts in Israel. 

Regarding classes, there are some variations in the numbers and definitions. BDD100K has 19 classes, whereas Cityscapes, KITTI and A2D2 have 30, 37 and 38 categories respectively. Fortunately, the benchmark evaluation that is done for Cityscapes excludes less prominent classes, and ends up being a set of 19 classes that exist in all four datasets. Thus, the only class modification we did is the naming convention such that all models interpret class calls in a similar fashion. 

In summary, the four datasets span four countries: Germany (34 cities), Switzerland (1 city), the United States (3 states) and Israel (8 districts). From these datasets, six evaluation sets are constructed as listed in Table~\ref{tb:data}. 

\begin{table}[]
\caption{Summary of the six evaluation sets. S, W, E, N refers to south, west, east and north, respectively.}
\label{tb:data}
\centering
\begin{tabular}{|lccl|}
\hline
\textbf{Dataset} & \textbf{Image dim} & \textbf{Images} & \textbf{Location} \\ \hline
Cityscapes Val & 2048x1024 & 500 & S, W and N Germany \\
KITTI Train & 1408x376 & 49 004 & Karlsruhe, Germany \\
KITTI Val & 1408x376 & 12 276 & Karlsruhe, Germany \\
BDD100K USA & 1280x720 & 3 281 & W and E coast USA \\
BDD100K Israel & 1280x720 & 362 & 8 districts in Israel \\
A2D2 & 1920x1208 & 41 277 & SE Germany \\ \hline
\end{tabular}
\end{table}

\subsection{Model selection}\label{sec:model}
The experiments in this paper utilize pre-trained models that have been trained on the Cityscapes training set without any adjustments to the original labels. From the Cityscapes leaderboard, three research architectures are selected, two from Deeplab v3+~\cite{chen2018encoder} and one from Pyramid Scene Parsing (PSPNet)~\cite{zhao2017pspnet}. All of the models are encoder-decoder style networks, where the encoder part of the networks receives features from a backbone that has been processed on the input image. The encoder then compresses the information into a lower-level representation. The decoder subsequently reconstructs feature masks of a higher-level representation based on the bottleneck of the encoder. % Finally, the decoder feature and bottleneck features are concatenated in a skip-connection, and processed through the model head where the semantic segmentation prediction is given. 

DeepLab-v3 provides two pre-trained versions with either ResNet101 (DLR) or Mobilenet-v2 (DLM) as the backbone. PSPNet also provides a pre-trained model with ResNet101 as the backbone. All models achieve similar mean intersection over union score (mIoU) on the Cityscapes validation set with 0.801, 0.809 and 0.825 mIoU for the DLM, DLR and PSPNet, respectively.

% The models differ in their encoder-decoder style, where DeepLabv3 uses atrous convolutions whereas PSPNet uses a pyramid pooling module combined with a spatial pyramid pooling module to extract features. 

% All models achieve ballpark similar mean intersection over union (mIoU) on the Cityscapes validation set with 0.801, 0.809 and 0.825 mIoU for the DLV3-MobileNet, DLV3-ResNet and PSPNet respectively.

% For the experiments, we test on two models. The first one is a pretrained DeepLab model trained solely on Cityscapes. The architecture consist of a MobileNet backbone that achieves a mIoU score of 0.721. 

% For training, a Feature Pyramid Network (FPN) is selected with a pre-trained ResNet34 feature extracting backbone. The backbone depth is five, which refers to amount of blocks that the backbone architecture is repeated. Each repetition reduces the spatial dimensions by 2, such that the dimensions are reduced by a factor 32 in spatial dimensions after the fifth block. The decoding part is repeating original papers' approach.

% The model is trained with Dice loss~\fix{CIT}{https://arxiv.org/pdf/1606.04797.pdf} combined with Adam optimization~\fix{https://arxiv.org/pdf/1412.6980.pdf}{} as both help reduce the risk to get stuck in local minima for data samples, where more seldom classes only exist in small regions. In addition, we compute the average intersection over union (IoU) with $>0.5$ overlapping ratio is considered as a true positive.   

\subsection{Evaluation metrics}\label{sec:metric}
This section presents the underlying equations used to express the OOD measure, risk, pixel coverage and experiment evaluation metrics. 

For outlier determination, we use the Mahalanobis Distance (MD) based on the prominent results from Lee et al.~\cite{Lee2018} where they introduced class-conditional Gaussian distributions as a basis for their MD. The MD measures the distance between a sample $X$ to a distribution $D$. The benefit of MD compared to Euclidean distance is that MD finds the eigenvectors representing the covariance of the distribution and thus, allows the distance metric to consider both the mean and covariance of the distribution when computing the distance. 

The class-conditional Gaussian distributions are accessed by extracting the true positive pixel subset of output vectors from training samples run through the DNN $f(\cdot)$. The subset is limited to $10^6$ pixels\textbf{} and then used to find the mean and covariance of the distribution $Q_{c} = \mathcal{N}(\mu_{c}, S_{c})$ for every class $c$ as 

\begin{equation}
\begin{split}
P_{c} &= P(f(\mathbf{x})|y=c) \\
\mu_{c} &= \frac{1}{N_{c}} \sum_{n=1}^{N_{c}} P_{c},\quad S_{c} = \frac{1}{N_{c}} \sum_{n=1}^{N_{c}} (P_{c} - \mu_{c})(P_{c} - \mu_{c})^{T}
\end{split}
\end{equation}

The class-conditional Gaussian distribution allows us to compute the MD as 

\begin{equation}
MD_{c}(o_{c}, Q_{c}) = \sqrt{(o_{c} - \mu_{c}) S_{c}^{-1} (o_{c} - \mu_{c})^{T} }
\label{eq:distance}
\end{equation}

where $o_{c} = P(f(\mathbf{x})|y=c)$ is the output mask of class $c$ from a model for a sample $\mathbf{x}$. Fig.~\ref{fig:intro_md_vis} shows a visualization of what the distance image may look like.

For risk, we define it as the opposite part of the IoU, i.e., considering the false positives and false negative parts of a prediction.  

Coverage is computed as the percentage of labeled pixels that received a prediction from the model, such that a prediction for every labeled pixel results in 100\% coverage. Note that a model prediction without a majority class is excluded by the model itself, hence the initial coverage can be less than 100\%. When extending the system with an accepted outlier threshold (MD in this paper), the coverage will be reduced and computed as the ratio between included pixels and the full image, see step 6 in Alg.~\ref{Alg:DistanceMetric}. 

By combining risk and coverage together with a distance metric, the risk-vs-coverage trade-off can be expressed as Alg.~\ref{Alg:DistanceMetric}. The algorithm provides risk as a function of the accepted threshold $\epsilon$, which similarly highlights pixel coverage as a percentage (0-100\%). Similarly, the area under the ROC-curve (AUC) is obtained by varying over the threshold $\epsilon$ with a discriminator as defined in Eq.~\ref{eq:discriminator} -- but instead looking at how the true positive rate and false positive rate vary. The AUC measure is also provided for each dataset, with respect to how the safety measure threshold adjusts the included pixels. 

From a safety engineering perspective, requirements can be put on the accepted level of risk of the Alg.~\ref{Alg:DistanceMetric} for a safety measure and thus, creating an optimization problem to find the needed discriminator threshold that maximizes coverage for the accepted risk. To emphasize this, two assumed safety requirements are formulated as 

\begin{itemize}
    \item[1.] The DNN shall not exceed 15\% risk. 
    \item[2.] The DNN shall maintain at least 50\% coverage. 
\end{itemize}

These requirements exemplify the usage of the safety measure and how it will indicate the performance of the model based on a given threshold.
% by reporting coverage when the risk is reduced to desired levels. 
% To this end, the coverage is reported when the risk is reduced to the desired target level. 
Note that the requirements are on the DNN only and not the full perception system that will have more strict safety requirements.   

\begin{algorithm}[!tpb] 
\caption{The risk-vs-coverage curve, yielded by varying the accepted distance $\epsilon$ of the OOD-metric.}
\label{Alg:DistanceMetric}
\begin{algorithmic}[1]
\Require DNN model $f(\cdot)$, OOD-method $\mathbf{M}(\cdot)$, varying threshold $\epsilon$, and label y. 
\State Compute the softmax model output \textbf{o} = f(\textbf{x}) for input sample \textbf{x} 
\State Let $c= argmax(\mathbf{o})$ be the predicted class $c$ and compute the distance score for class output vector $\mathbf{o}_{c}$ with $\mathbf{M}(\mathbf{o}_{c})$.  
\For{$\epsilon$ in $\{0, \epsilon_{1}, \epsilon_{2}, ..., 1 \}$}    
\State Let discriminator $D\{0, 1\}$ be defined as \begin{equation}\label{eq:discriminator}
    D = \left\{\begin{matrix}
\begin{matrix}
1 & if\quad \textbf{M}(\textbf{o}_{c}) <  \epsilon \\ 
0 & otherwise
\end{matrix}
\end{matrix}\right.
\end{equation}
\State Let p = $\mathbf{o}(D=1)$ be the accepted pixel subset. 
\State Compute prediction risk and pixel coverage as 
\begin{align*}
    Risk &= 1 - IoU(p)\\
    cov &= p / y
\end{align*}
\EndFor
\end{algorithmic}
\end{algorithm}

\subsection{Evaluation technique}\label{sec:eval}
Tying together Sections~\ref{sec:data} and \ref{sec:metric}, the objective of this paper is to answer RQ1 through average discrimination performance through the AUC measurement, with the hypothesis that the score is decreased on datasets with different sensor positions, geolocations or labeling approaches. All datasets except Cityscapes are considered to be part of the outlier sets, as they are either gathered in a different country, contain a different camera setup, or were annotated using different labeling guidelines. 
%When comparing trained data to outlier data that is far off from the training set, and RQ2 to be answered with the graph constructed from Alg.~\ref{Alg:DistanceMetric} where the hypothesis is that risk being reduced by limiting coverage. 

% For inlier and outlier sets, we consider the BDD100K subsets (i.e., Test 2 and 3 in Table~\ref{tb:datasplit}) as an outlier set, as it is gathered in foreign country, as well as a different camera setup compared to the Cityscapes dataset. 

For RQ2, the risk-coverage curves visualize the trade-off between misclassifications and amount of pixels predicted. The hypothesis is that the risk and coverage will be reduced as the restrictiveness of the safety measure increases, i.e., as the accepted MD for a pixel is reduced, the safety measure rejects more samples but manages to prioritize the removal of pixels that are more likely to be misclassified. If there exists a rejection threshold that fulfils the assumed safety requirements, the evaluation set is considered in-distribution.

%%%%%%%%%%%%%%%%%%%%%%%%%%%%%%%%%%%%%%%%%%
%%%%%%%%%%%%%%%%%%%%%%%%%%%%%%%%%%%%%%%%%%
\section{Results}\label{sec:results}
The results section is divided into two parts: The first one aims at reviewing the algorithmic results when applying the safety measure on the different evaluation sets, and the second part evaluates if the usage of the safety measure contributes to the safety argumentation. 

\subsection{Metrics evaluation}\label{sec:results-eval}
\begin{figure}[t]
\centering
\includegraphics[width=0.99\linewidth]{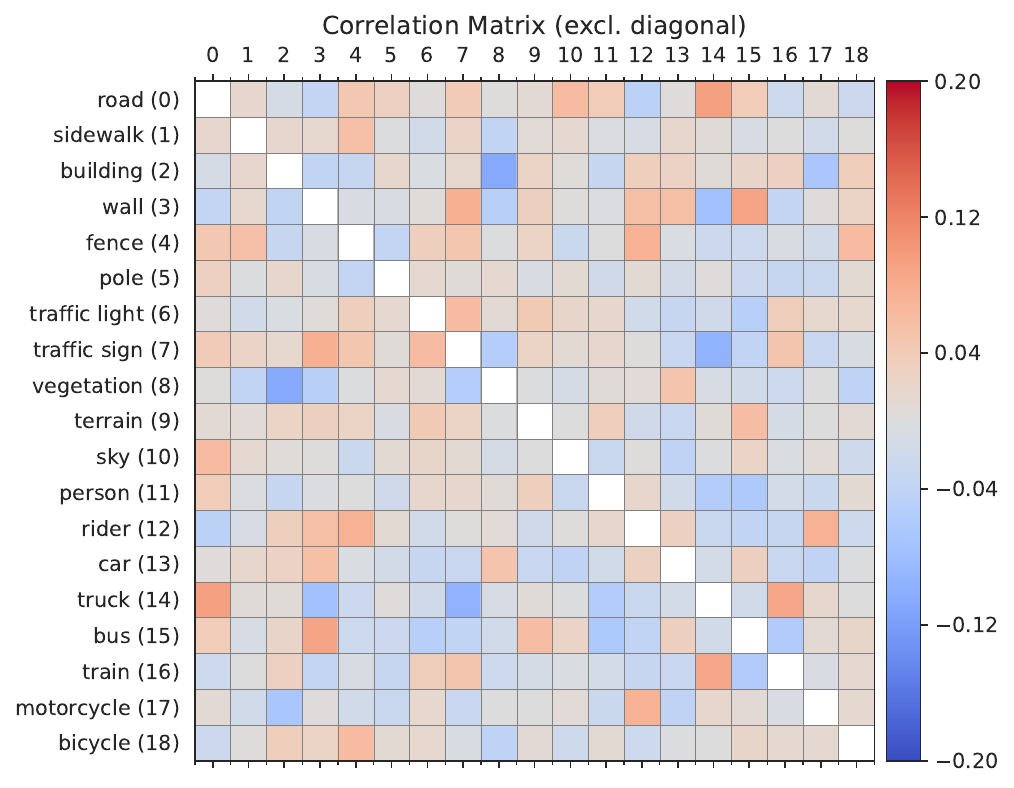} 
\caption{Correlation between classes for the PSPNet model. Note that the diagonal is excluded, as correlation with oneself is always 1.}
\label{fig:class_correlation}
\end{figure}

The initial steps of the evaluation constitute extracting the class conditional Gaussian distribution for the three models. The distribution is extracted from the Cityscapes training samples and is constructed to gather up to a million pixels per class with the limitation of a maximum of ten thousand pixels for one class from one individual image. From the set of pixels, the mean covariance matrices are computed, resulting in $\boldsymbol{\mu}$ a 19$\times$19 matrix of means, where $\boldsymbol{\mu_i}$ corresponds to the mean vector of class $i$ and covariance matrices $\boldsymbol{S}$ a 19$\times$19$\times$19 matrix, such that $\boldsymbol{S_i}$ corresponds to the covariance matrix for correctly classified elements of class \textit{i}. Visualizations of the covariance matrices for all three models can be seen with the evaluation code\footnote{code available at: \href{https://github.com/jenshenriksson/ood-ad-comparison}{https://github.com/jenshenriksson/ood-ad-comparison}}. In this paper however, for a straightforward comparison, the correlations between classes are evaluated and presented in Fig.~\ref{fig:class_correlation}. In the plot, the Pearson correlation coefficient is visualized (with the exception of the diagonal, which is always one) for the PSPNet model, referring to a measure of the linear relationship between classes in the dataset.

% After the retrieval of city locations based on the reversion of GPS coordinates described in Section~\ref{sec:data}, the four datasets contribute to three test sets: 

% \textbf{Test set 1} is constructed as an inlier test set from Cityscapes. As the labels for the original test set is hidden for the Cityscapes benchmark suite, we construct an inlier test set of three cities from which the risk-coverage ratio can be analyzed. The cities are Bremen, Hamburg and Hanover, and together test set 1 covers 760 images. 

% \textbf{Test set 2:} is the subset of images from BDD100K with images from Israel. The reverse location are grouped based on districts. All six main administrative districts of Israel are included, and in total it contains 362 images. 

% \textbf{Test set 3} is the BDD100K subset of images that was gathered in California, US. In total it covers 491 images across the state. 

% \textbf{Test set 4} is one sequence from A2D2. The reversion yielded four main cities in Germany, namely München, Nürnberg, Ingolstadt and Regensburg with an additional 8 suburbs to the cities. The selected run has identifier ID \verb20180807_145028 that corresponds to a run in the city Ingolstadt with 942 images. 

The mean and inversed covariance are used to compute the distance measure from Eq.~\ref{eq:distance}. During the evaluation, we evaluate the rejection rate based on individual distances, i.e., the threshold is iterated between $MD_{min}$ to $MD_{max}$ of the sample image with $60$ threshold points. This is infeasible for online usage, however by allowing each image individual evaluation, it yields the safety measures' optimal performance. For the graphs however, the resulting risk and coverage are averaged out per threshold point per dataset.  

The results for the experiments are visualized in risk-coverage graphs in Fig.~\ref{fig:risk-coverage} (reviewed in the next Section), and key performance measures are presented in Table~\ref{tb:results}. Starting off, studying the baseline performance on the Cityscapes datasets. The three models, when running on the training set, yield 86.21\%, 84.81\% and 85.53\% IoU for PSPNet, DLM and DLR, respectively. All models have similar AUC of $0.91-0.92$, showing good separability as well as all models maintain close to 100\% coverage as the inherent risk is below the accepted risk levels defined in Section~\ref{sec:eval}. The small discrepancy in coverage is due to the models' inherent rejection possibility that occurs when there is no prediction with a majority, hence the network abstains from giving any prediction. On the validation set, the IoU scores are reduced to 82.48\%, 80.05\% and 80.95\% for PSPNet, DLM and DLR, respectively. The performance drops by an average 4.36\% IoU points, which can be considered high as the validation set represents an unbiased evaluation of the models' performance on previously unseen data. However, the phenomenon occurs in all three models, which can instead indicate that the validation set is slightly more challenging than the training set. For both Cityscapes sets, PSPNet shows marginally better performance in IoU, thus is the better choice of model.

BDD100K constitutes the next evaluation set. Starting with the parts from the USA, the IoU scores see a dramatic downturn to 46.86\%, 47.57\% and 57.31\% and the AUC scores drop to 0.63, 0.61 and 0.73 for PSPNet, DLM and DLR, respectively. It is evident that the IoU performances of all three models are significantly lower on this BDD100K evaluation set compared to the Cityscapes sets, with a subsequent reduction in AUC scores. The difference in performance may be attributed to different dataset properties, i.e., differences in scene composition, object formation and camera properties. Regarding camera properties, it is noted that for BDD100K, not all images are forward-looking, e.g., the camera angle difference shown in Fig.~\ref{fig:test_set_images} for BDD100K (second image from the left) compared to the remaining test images. The same results are seen for the Israel part of BDD100K. The model's IoU is reduced slightly more to 45.07\%, 42.01\% and 57.31\% for PSPNet, DLM and DLR, respectively. For DLM, the change is significantly lower compared to the USA set, with a reduction of 5.56\% IoU units, whereas the other models are only marginally lower. For both BDD100K sets, the DLR shows significantly better IoU scores as well as AUC, thus being the better choice of model. PSPNet was originally better, based on Cityscapes evaluation, but seems less able to manage samples far off the training domain.

\begin{table*}[]
\centering
\caption{The results from running the MD-evaluation with 60 threshold points on the training set and the six evaluation sets. Upward arrows ($\uparrow$) indicate higher values are better. Check marks ($\checkmark$) indicate if the safety requirements are fulfilled. }
\label{tb:results}

\begin{tabular}{|l|ccc|ccc|ccc|}
\hline
\textbf{Dataset} & \multicolumn{3}{c|}{\textbf{PSPNet}} & \multicolumn{3}{c|}{\textbf{DLM}} & \multicolumn{3}{c|}{\textbf{DLR}} \\ \hline
 & \textbf{IoU (\%) $\uparrow$} & \textbf{AUC $\uparrow$} & \textbf{FS1 cov (\%) \checkmark} & \textbf{IoU (\%) $\uparrow$} & \textbf{AUC $\uparrow$} & \textbf{FS1 cov (\%) \checkmark} &\textbf{IoU (\%) $\uparrow$} & \textbf{AUC $\uparrow$} & \textbf{FS1 cov (\%) \checkmark}\\ \hline
\textbf{Cityscapes Train}	&	86.21	&	0.92	&	99.48 \checkmark	&	84.81	&	0.92	&	99.50 \checkmark	&	85.53	&	0.91	&	99.66 \checkmark\\
\textbf{Cityscapes Val}	&	82.48	&	0.89	&	99.19 \checkmark	&	80.05	&	0.88	&	98.18 \checkmark	&	80.95	&	0.88	&	98.67 \checkmark\\
\textbf{BDD100K USA	}&	46.86	&	0.63	&	0.08	&	47.57	&	0.61	&	0.06	&	57.45	&	0.73	&	0.61 \\
\textbf{BDD100K Israel}	&	45.07	&	0.61	&	0.16	&	42.01	&	0.54	&	0.01	&	57.31	&	0.69	&	0.07 \\
\textbf{KITTI Train}	&	74.41	&	0.77	&	94.59 \checkmark	&	71.66	&	0.73	&	62.22 \checkmark	&	76.10	&	0.80	&	98.31 \checkmark\\
\textbf{KITTI Val}	&	72.81	&	0.76	&	86.19 \checkmark	&	70.92	&	0.72	&	46.36	&	74.73	&	0.79	&	93.48 \checkmark\\
\textbf{A2D2}	&	59.38	&	0.64	&	0.27	&	52.82	&	0.61	&	0.01	&	68.77	&	0.69	&	42.66 \\ \hline

\end{tabular}
\end{table*}

For the KITTI training set, the IoU values were 74.41\%, 71.66\% and 76.10\% and the AUC values were 0.77, 0.73, and 0.80 for PSPNet, DLM, and DLR, respectively. DLR outperforms the remaining models in all regards for the KITTI training set. The scores are more in line with Cityscapes, approximately 5-10\% lower than those in the Cityscapes validation set for all three models, but still exhibit some influence from differences in dataset properties. The one clear difference we found in KITTI is that the camera dimensions are wider (3.74:1, width to height ratio) compared to Cityscapes' 2:1 ratio. Otherwise, the dataset definitions are the same, as KITTI annotation style is based on Cityscapes. For the KITTI validation set, minor, yet consistent reductions in IoU are seen for all models. However, DLR still outperforms the two other models with IoU score of 74.73\% compared to 72.81 and 70.92 for PSPNet and DLM, respectively.  

On the last evaluation set from Audi the results are once again on a downturn. The IoU scores for the models were 59.38\%, 52.82\% and 68.77\% and the separability performance resulted in AUC of 0.64, 0.61 and 0.69 for PSPNet, DLM and DLR, respectively. PSPNet demonstrated better performance than DLM with both higher IoU and AUC, but still has considerably lower performance than DLR. Surprisingly, the results on the A2D2 dataset are far off compared to KITTI and Cityscapes, even though the dataset shares several similarities in dataset layout, image dimensions and direction of the forward-looking camera. One discrepancy found during experimentation is that A2D2 has divided some broader classes into more detailed classes, which may contribute to additional false positive activations in the models. 

Summarizing the tabular results, discrepancies in labels and camera position have a major impact in determining performance on unseen data. All models achieved good results on the Cityscapes validation set, which consists of new cities in proximity to the training cities. Acceptable results are achieved on the KITTI sets, especially for PSPNet and DLR models that managed to pass the assumed safety requirements. KITTI is gathered in Karlsruhe, Germany, which also is in proximity to the annotated cities in Cityscapes. However, for BDD100K and A2D2 all models fall short, and do not manage to show generalizability.

% However, the evaluation sets from Berkley Deep Drive are considered OOD as all models are unavailable to achieve adequate results on the sets. Finally, A2D2 is also considered OOD as safet

% The first research question is analyzed as the summation of all distance metrics in relation to all classified pixels, see Eq.~\ref{eq:sum_of_MD}. The reasoning behind this is if there's is any separability between datasets, there has to be an increase in uncertainty when an uncertain scenario is occurring. We initiate this study by comparing test set 1 with test set 2, i.e., scenery from German (inlier) cities compared to images from Israel (outlier), see Fig.~\ref{fig:israel-vs-germany}. Investigating the graph, one can see there is no separability between the inlier and outlier sets with this form of measure. In fact, even some inlier samples score higher outlier scores than samples from the decided outlier set. Repeating the experiment shows the same results when comparing the inlier set to test set 3 and 4. Keep in mind, test set 1 represents data that the model has previously seen, which gives the indication that the model generalizes well enough to not give special considerations towards the set, but instead shows similar behavior no matter where the data has been gathered. Intuitively this is to be expected for classes that are of similar nature around the world, but surprisingly it also is the case for classes that deviate depending on region, e.g., road signs that are of different shape and alphabet depending on where the sample has been gathered.

\subsection{Applicability to Safety Requirements}\label{sec:results-safety}
\begin{figure*}[!t]
    \centering
    \includegraphics[width=0.99\linewidth]{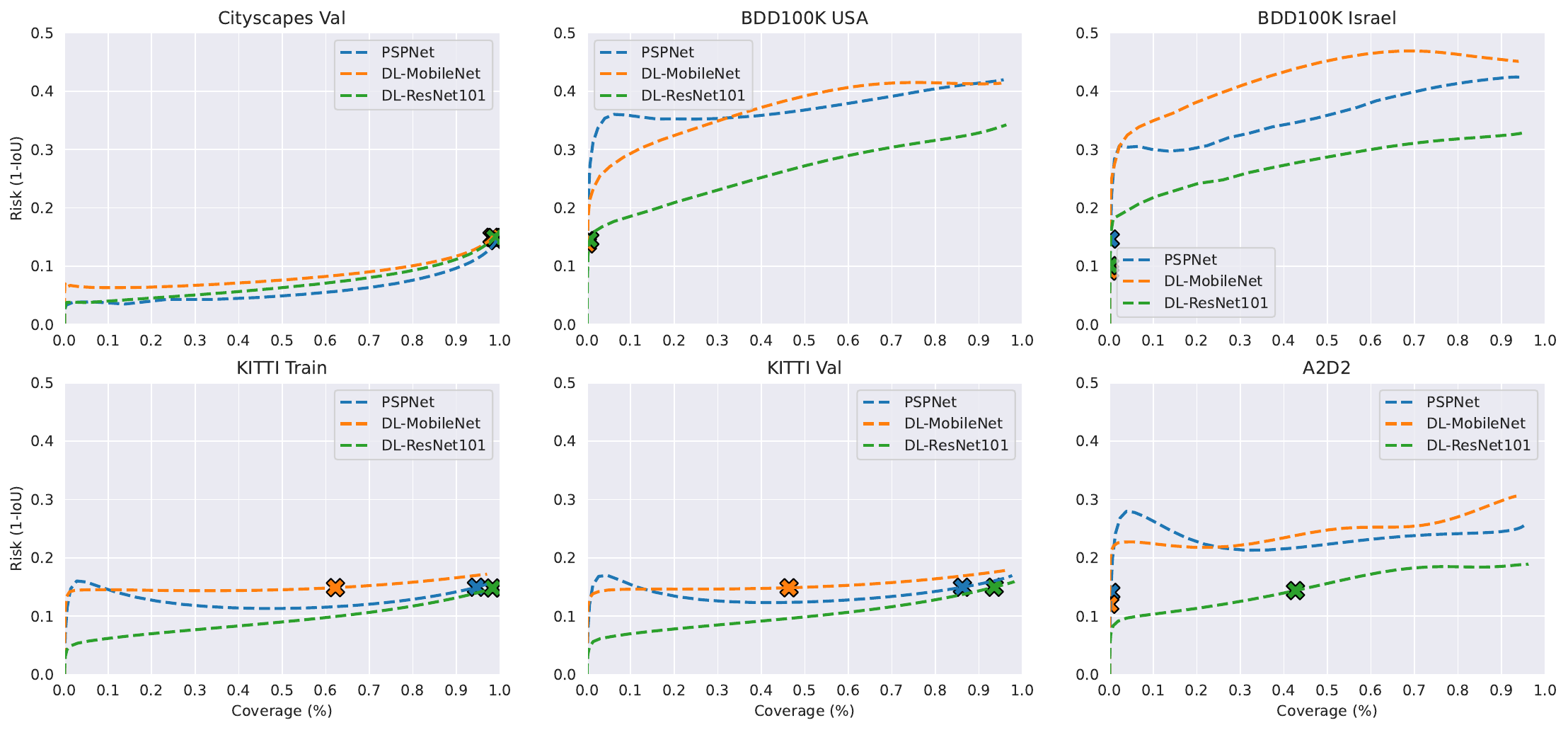}
    \caption{The risk-coverage showcasing the trade-off plots for the six evaluation sets. The cross-markers (\ding{54}) visualize the breakpoints where the assumed risk requirement is fulfilled for each of the trained models per evaluation set. 
    %Similar behavior is seen in all except the last one, where risk is first increased before showcasing the risk reduction.
    }
    \label{fig:risk-coverage}
\end{figure*}

The threshold variations of the experiments are visualized in Fig.~\ref{fig:risk-coverage}, where the resulting risk and coverage are plotted as described in Alg.~\ref{Alg:DistanceMetric}. The safety requirement is formulated as a hypothetical target of minimum 50\% coverage with 15\% classification risk, which determines if a model is able to operate in a region that is outside of the training domain. The intersection of the risk-curve and accepted risk level is marked with a cross in the graph for each model. For the assumed requirement, all models pass on Cityscapes training and validation set, and partly succeed on KITTI, where only DLM is borderline accepted on the training set but falls short on the validation set. 

For the BDD100K evaluation sets, none of the models pass the safety requirements. In fact, the accepted risk level would need to be increased to 28.59\% to achieve a 50\% coverage rate for the best model DLR. In a similar fashion, DLR just barely falls short on the A2D2 evaluation set. For this evaluation set, the risk elicitation requires a slight increase to 15.24\%. Without any elicitation, both BDD100K and A2D2 evaluation sets are considered OOD.

The expectations of the risk-coverage curves are a monotonic decrease in risk when increasing the strictness of the safety measure. Studying the graphs, this is not evident based on the results as, e.g., PSPNet shows on several evaluation sets that the risk is in fact increased with higher restrictiveness. This is not covered in any of the safety requirements, but could be extended to be put as a requirement, as risk increments indicate overfitting, as the model is overconfident and removes true positives rather than false positives.

% \begin{equation}
% \label{eq:sum_of_MD}
%     \sum_{p}^{} MD(p | \mathbf{o}_{c} \ne 0)
% \end{equation}

%  Breaking down Fig.\ref{fig:israel-vs-germany}, we discovered that all classes, no matter the number of samples included in the training set, will yield similar Mahalanobis distances for a well-trained net. This is seen by studying the variation of outlier scores for the different classes when compared to each other. We find that the majority of outlier metrics are of similar Gaussian distribution. 

% \review{\lipsum[1-5]}

% \review{\lipsum[1-6]}
%%%%%%%%%%%%%%%%%%%%%%%%%%%%%%%%%%%%%%%%%%
%%%%%%%%%%%%%%%%%%%%%%%%%%%%%%%%%%%%%%%%%%
\section{Discussion}\label{sec:fw}
OOD detection is a critical task for automotive perception. The community acknowledges that there will be no complete formal description of what constitutes the perception of specific objects, e.g., pedestrian detection.  Instead, the challenge will be to maximize the normative performance of the system, and to incorporate safety measures that reduce the probability of false positives in the system. 

This paper shows how to apply a safety measure for semantic segmentation deep nets on autonomous driving datasets. By studying the varying performance of the model while performing inference on input samples that are diverse compared to the training data, safety evaluators get an indication of how well the model will perform on samples outside of the training domain. The best dataset comparison achieved in this paper is the Cityscapes-KITTI evaluation. All models sustain performance with minor decreases when evaluated on the validation set from Cityscapes. However, on every other evaluation set, the performance is drastically decreased. On KITTI, the models still maintain acceptable results and manage to fulfill the hypothetical requirements. In KITTI-360~\cite{Liao2022KITTI-360:3D}, they explicitly state their labeling process builds upon Cityscapes, thus the label extension can be neglected and the remaining classes are overlapping. 

However, overlapping labels do not infer data being in-distribution. The semantic segmentation labels from BDD100K~\cite{Yu2020BDD100K:Learning} are fully overlapping with the labels used for evaluation in Cityscapes benchmark suite~\cite{Cordts2016TheUnderstanding}, but evaluation performance is underwhelming on both evaluation sets from BDD100K. For A2D2, the labeling convention is slightly more detailed, something we neglected in these experiments, but seems to affect the results negatively more significantly than expected. One solution for this would be to merge some labels into a broader class (e.g., merging the classes \verb|drivable cobblestone| with \verb|RD normal street| in A2D2). In summary, consistency of class definitions, labeling methodology and sensor setups are key to being able to compare between evaluation sets.

While our evaluation concluded that BDD100K and A2D2 are considered OOD, a better solution would be an iterative process with the aim to achieve a verified perception system by identifying the weak points of the system as a whole and breaking them down into sub-parts, each corresponding to a specific part of the ODD. To this end, an extension of test and training scenarios with corresponding data can be constructed where the goal of the iterative process is to either improve the performance of the system, or with the help of safety measures to highlight where the model is out of scope. If neither is possible, the scenario is considered OOD and testing shows that instead a limitation of the functionality is needed, i.e., a restriction in the desired ODD to ensure that the performance of the functionality meets its requirements. While this paper does not conduct this iterative process, it indicates that the process as a whole is feasible. By the design of the risk-coverage trade-off, the machine learning field is able to formulate the varying risk depending on how restrictive the safety engineer deems the system to be. 
It is noteworthy, that solely rejecting a prediction does not remove the potential risk-- rather it highlights that the prediction is uncertain and the vehicle should rather proceed with caution, rather than continue as before. 

% Unfortunately, this paper could not highlight any separability between an inlier set and an outlier set, even when collecting data from multiple independent datasets with varying gathering techniques. One could however expect this phenomenon, as the dimensional increase is several magnitudes compared to the datasets that have been used in previous experiments. For example, Yang et al.~\cite{YangOpenOOD:Detection}

\subsection{Threats to validity}
The authors acknowledge that the datasets differ in sensor equipment, scene and object composition, and image quality, thus providing an unfair comparison, as the models are only trained on one dataset. Nevertheless, the same performance variations are seen in independent models. Furthermore, the resulting MD method is not the sole safety measure, but instead should be part of an iterative verification process, where this is one out of many measures that improve the quality of the deep learning prediction.

\section{Conclusions}
This paper has shown that the risk-coverage trade-off exists for pixel coverage just as in deep learning classification tasks. An evaluation set can be considered OOD as a whole, but difficulties still exist on a per-image basis. Our study shows that risk can be reduced by only accepting predictions above an accepted distance threshold. We show this phenomenon with Mahalanobis distance as a safety measure across four AD datasets that together span different styles of driving scenarios. Furthermore, we find that discrepancies in dataset properties impact the performance drastically, and suggest future experiments re-train models with images from different datasets for a more fair comparison. 

\clearpage
\section*{Acknowledgments}
This research has been supported by the Strategic vehicle research and innovation (FFI) programme in Sweden, via the project SALIENCE4CAV (ref. 2020-02946) and has been supported by the Wallenberg AI, Autonomous Systems and Software Program (WASP) funded by the Knut and Alice Wallenberg Foundation.

\bibliographystyle{IEEEtran}
\bibliography{Mendeley,Extra}

\end{document}